\documentclass[10pt,twocolumn,letterpaper]{article}

\usepackage[pagenumbers]{cvpr} 

\usepackage{graphicx}
\usepackage{amsmath}
\usepackage{amssymb}
\usepackage{booktabs}

\usepackage[pagebackref,breaklinks,colorlinks]{hyperref}

\usepackage[capitalize]{cleveref}
\crefname{section}{Sec.}{Secs.}
\Crefname{section}{Section}{Sections}
\Crefname{table}{Table}{Tables}
\crefname{table}{Tab.}{Tabs.}

\def\cvprPaperID{9005} 
\def\confName{CVPR}
\def\confYear{2022}

\usepackage[accsupp]{axessibility}  

\usepackage{overpic}
\usepackage{enumitem} 
\usepackage{overpic} 
\usepackage{color}
\usepackage{float} 
\usepackage{placeins} 

\usepackage{mathtools} 

\definecolor{turquoise}{cmyk}{0.65,0,0.1,0.3}
\definecolor{purple}{rgb}{0.65,0,0.65}
\definecolor{dark_green}{rgb}{0, 0.5, 0}
\definecolor{orange}{rgb}{0.8, 0.6, 0.2}
\definecolor{red}{rgb}{0.8, 0.2, 0.2}
\definecolor{darkred}{rgb}{0.6, 0.1, 0.05}
\definecolor{blueish}{rgb}{0.0, 0.3, .6}
\definecolor{light_gray}{rgb}{0.7, 0.7, .7}
\definecolor{pink}{rgb}{1, 0, 1}
\definecolor{greyblue}{rgb}{0.25, 0.25, 1}

\newcommand{\todo}[1]{{\color{red}#1}}
\newcommand{\TODO}[1]{\textbf{\color{red}[TODO: #1]}}

\newcommand{\RR}[1]{{\color{dark_green}{\bf [RR: #1]}}} 

\newcommand{\TA}[1]{{\color{blueish}{\bf [TA: #1]}}}

\newcommand{\loss}[1]{\mathcal{L}_\text{#1}}

\newcommand{\Fig}[1]{Fig.~\ref{fig:#1}}
\newcommand{\Figure}[1]{Figure~\ref{fig:#1}}

\newcommand{\Table}[1]{Table~\ref{tab:#1}}

\newcommand{\Equation}[1]{Equation~\ref{eq:#1}}
\newcommand{\Sec}[1]{Sec.~\ref{sec:#1}}

\usepackage{blindtext}

\renewcommand{\paragraph}[1]{\vspace{1mm}\noindent\textbf{#1}.} 

\usepackage{pifont}
\newcommand{\cmark}{\ding{51}}%
\newcommand{\xmark}{\ding{55}}%

\newcommand{\figcaption}[2]{\caption{\textbf{#1} #2}}

\newcommand\blfootnote[1]{%
  \begingroup
  \renewcommand\thefootnote{}\footnote{#1}%
  \addtocounter{footnote}{-1}%
  \endgroup
}

\definecolor{cone}{RGB}{31,119,180}
\definecolor{ctwo}{RGB}{255,127,14}
\definecolor{cthree}{RGB}{44,160,44}

\usepackage{ifthen}
\newcommand{\final}{1}  
\ifthenelse{\equal{\final}{1}}
{
    \renewcommand{\TA}[1]{}
    \renewcommand{\RR}[1]{}
    \renewcommand{\todo}[1]{}
    \renewcommand{\TODO}[1]{}
}{}
\begin{document}
\title{NPBG++: Accelerating Neural Point-Based Graphics}

\author{
Ruslan Rakhimov\textsuperscript{1}\thanks{Equal contribution}\quad Andrei-Timotei Ardelean\textsuperscript{1}\footnotemark[1]\quad Victor Lempitsky$^{1,2}$\quad Evgeny Burnaev$^{1,3}$\\\\
\textsuperscript{1}Skolkovo Institute of Science and Technology, Russia\\
\textsuperscript{2}Yandex, Russia\quad\textsuperscript{3}Artificial Intelligence Research Institute, Russia\\
}
\maketitle
\begin{abstract}
We present a new system (NPBG++) for the novel view synthesis (NVS) task that achieves high rendering realism with low scene fitting time.  Our method efficiently leverages the multiview observations and the point cloud of a static scene to predict a neural descriptor for each point, improving upon the pipeline of Neural Point-Based Graphics \cite{aliev2020neural} in several important ways. By predicting the descriptors with a single pass through the source images, we lift the requirement of per-scene optimization while also making the neural descriptors view-dependent and more suitable for scenes with strong non-Lambertian effects. In our comparisons, the proposed system outperforms previous NVS approaches in terms of fitting and rendering runtimes while producing images of similar quality. Project page: \url{https://rakhimovv.github.io/npbgpp/}.
\end{abstract}
\section{Introduction}
\label{sec:intro}



\blfootnote{Correspondence to {\tt\footnotesize ruslan.rakhimov@skoltech.ru}}

\begin{figure}[t]
\centering
\includegraphics[width=0.99\columnwidth]{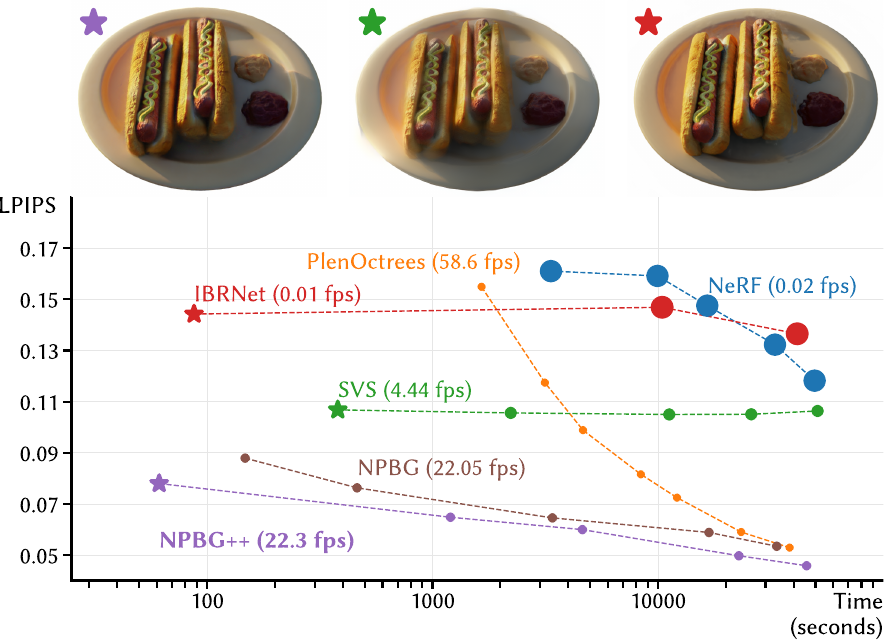}
\figcaption{Time comparison.}{
Time vs. image quality (LPIPS) comparison of several methods, computed on the `hotdog` scene from the NeRF-synthetic dataset. The time axis represents the time-to-rendering, i.e.,~fitting time + rendering time for one image. For methods marked with $\star$, the first scores are reported without per-scene optimizations. Fitting time consists of feature extraction for IBRNet, geometry estimation + 3D modeling stage for NPBG++, and geometry estimation + meshing for SVS (the renderings on top offer qualitative comparisons between these configurations). The remaining scores are computed at different points in the fine-tuning processes. Circle areas are proportional to logarithms of the rendering times (smaller is better) and highlight the methods’ rendering speed.
}
\label{fig:teaser}
\end{figure}


The ability to render photorealistic views of a scene, as learned from a handful of observations, opens the door to many applications in virtual / augmented reality, cinematography, gaming industry, and virtually any field which involves computer graphics. While there is a high interest in creating such novel view synthesis (NVS) systems, the problem proves to be challenging. Early methods based on view interpolation \cite{chen1993view, kanade1995virtualized, mcmillan1995head} do not fare well enough in real-world scenarios, which entail complex geometry, limited proximity of input images, lighting variations, etc. There are mainly three development directions which the different works addressing this task generally seek to improve: rendered image quality, scene fitting time, and rendering speed. Despite the recent development in Computer Vision brought by Deep Learning approaches, there is still a noticeable gap between the current state of the art in NVS and an ideal model for this task.

To this end, our work proposes a new system that enables real-time rendering and can rapidly adapt to new scenes. Similarly to \textit{Neural Point-Based Graphics} (NPBG)~\cite{aliev2020neural}, our approach uses point clouds to model the geometry of scenes. This representation is advantageous as point clouds are generally available by using inexpensive RGBD cameras or by processing RGB streams with classic Structure-from-Motion and Multi-View Stereo pipelines such as COLMAP~\cite{schoenberger2016sfm, schoenberger2016mvs}.
These reconstructions are usually not accurate or complete enough to be used directly for rendering, whereas performing surface estimation could yield a loss of geometric details and requires significant additional computations. Instead, we devise our method to work directly on raw point geometry and address the problem of small noise and low point density using neural rendering.

Using a point-based geometry together with a neural rendering model had been shown to yield good results as NVS systems \cite{aliev2020neural, kolos2020transpr, kopanas2021point, lassner2021pulsar}. However, these approaches require per scene optimization of per-point descriptors and optionally a deep rendering network, resulting in a lengthy process to achieve high-quality renderings. To accelerate this process, our method predicts the points' features from the source images, enabling fast scene representations, which can then be rendered in real-time. If needed, one can further finetune these predictions to increase the quality of the result. The challenge of the approach is the proper integration of data from several views while considering view-dependent appearance, occlusions, and missing information.

The system we propose is effective and fast, see \Figure{teaser}: for each input image and associated camera parameters, we run the feature extractor network on it, yielding feature maps representing each pixel's local appearance. The point cloud is then projected onto the image, taking into account occlusions to obtain features. An online aggregation method was devised to effectively aggregate the obtained features from one image at a time. After processing all input views, we get the final view-dependent neural descriptors for each point. This representation is then rendered by a U-Net shaped network that integrates multi-scale rasterizations of the point cloud similarly to NPBG \cite{aliev2020neural}.

To further improve the quality of the NPBG system, we introduce two important modifications (that can also be applied in analogous systems). Firstly, we show how view-dependency can be added to neural descriptors efficiently, without an excessive increase of scene fitting time, while boosting the quality of NPBG for scenes with non-Lambertian surfaces. Secondly, we introduce two lightweight 2D alignment stages in the pipeline that address the non-equivariance of convolutional parts of the pipeline to in-plane rotations.

To summarize, our main contributions are:
\begin{itemize}
    \item a new NVS system capable of quickly generating neural scene representations that can then be rendered at interactive rates at high resolution.
    \item an online, permutation-invariant, aggregation method for incorporating features from any number of source views using constant memory into neural descriptors, which facilitates view-dependent effect modeling.
    \item an alignment technique that makes the rendering process equivariant to in-plane rotations, appropriate for any pipeline working with neural descriptors estimated from images.
\end{itemize}
\section{Related work}
\label{sec:related}


\paragraph{Novel view synthesis} The computer vision and graphics communities have long been interested in the problem of novel view synthesis and, specifically, image-based rendering. Several approaches have been developed that differ in how they leverage and represent the underlying geometry of the scenes. Ranging from methods like light field rendering~\cite{levoy1996light} that do not rely on an explicit geometric proxy and require a dense set of image samples instead~\cite{shum2000review}, to methods that rely on accurate geometry for synthesis. More recently, the interest amplified around the idea of applying deep learning models to replace or enhance parts of the classic pipelines. Among others, neural networks were employed to predict blending weights for composing input images~\cite{hedman2018deep}, predict camera poses~\cite{zhou2017unsupervised}, depth maps~\cite{kalantari2016learning}, multi-plane images~\cite{flynn2016deepstereo, zhou2018stereo}, and voxel grids~\cite{NIPS2017_9c838d2e}. Moreover, several recent methods harness neural scene representations to build upon standard 3D scene reconstructions using different proxy geometries such as point clouds~\cite{aliev2020neural, meshry2019neural, pittaluga2019revealing} or meshes~\cite{thies2019deferred, riegler2020free, riegler2021stable}. 
An important breakthrough was brought by the introduction of Neural Radiance Fields (NeRF)~\cite{mildenhall2020nerf} where it is proposed to model the entire scene continuously using a fully connected network that is optimized using differentiable volume rendering. The approach has lately seen many variants which address some of its limitation: the large number of required views~\cite{chibane2021stereo, yu2021pixelnerf}, inter-observations variance~\cite{martin2021nerf}, rendering speed~\cite{liu2020neural, neff2021donerf, garbin2021fastnerf, yu2021plenoctrees} and scene fitting time~\cite{chen2021mvsnerf, chibane2021stereo, yu2021pixelnerf, tancik2021learned}.

NVS systems can be generally divided into two categories: methods that require timely per-scene optimization~\cite{aliev2020neural, mildenhall2020nerf, wizadwongsa2021nex} and methods that can easily and rapidly generalize to new scenes by learning to integrate information from the input views efficiently~\cite{chen2021mvsnerf, chibane2021stereo, riegler2021stable, yu2021pixelnerf, wang2021ibrnet}. Most systems in the second category also allow fine-tuning for particular scenes making them more versatile and preferable to the first group. Our proposed system, NPBG++, belongs to the latter category, as it can aggregate information from all available observations in a single pass.

\begin{figure*}[ht]
\hsize=\textwidth
\centering
\includegraphics[width=0.99\textwidth]{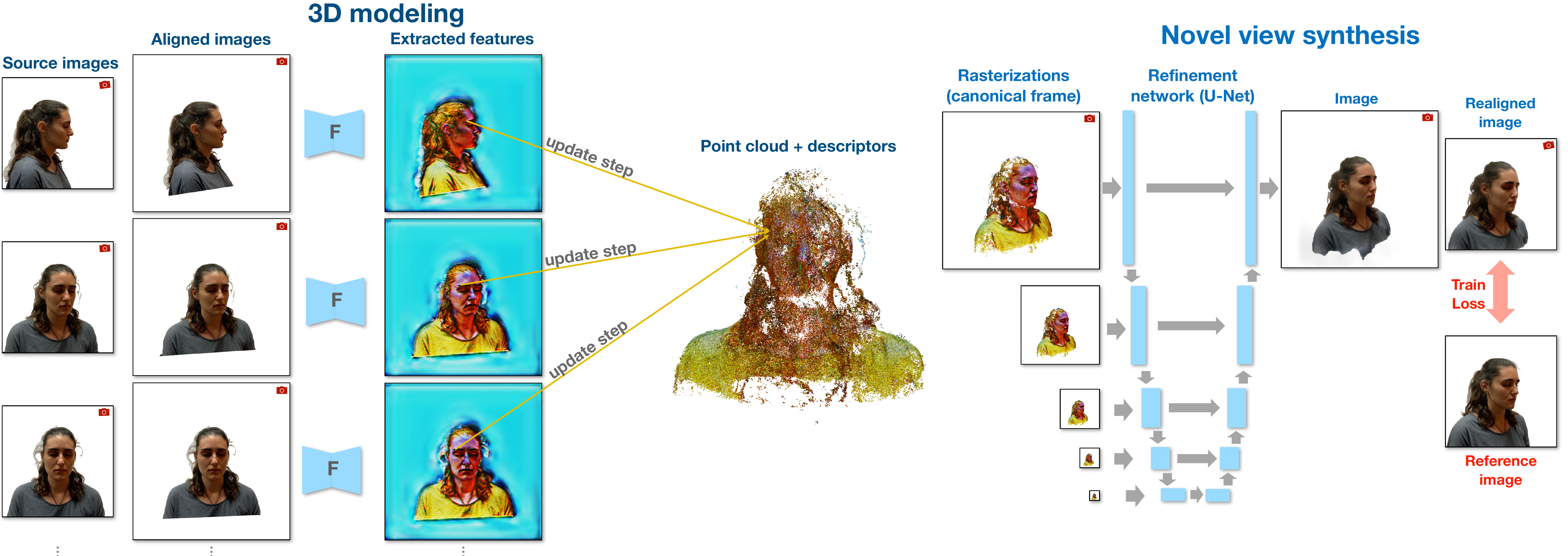}
\figcaption{Overview of NPBG++: the system for fast novel view synthesis.}{We represent the scene as a point cloud with a view-dependent neural descriptor embedded in each point. During 3d modeling stage (\Sec{preparation_stage}), we sequentially process each input view (input image alignment and feature extraction) and apply online aggregation to update the neural descriptors of each point (no fitting). During the novel view synthesis stage (\Sec{test_stage}), we rasterize the point cloud, pass the rasterization result through the rendering network, and post-process it (output image alignment) to get the novel view.}
\label{fig:overview}
\end{figure*}


\paragraph{Point-based graphics}
The use of points as a rendering primitive was identified early on~\cite{csuri1979towards, levoy1985use} and sparked interest due to the simplicity and efficiency it conveys. While they are fast and have little redundancy, point clouds have been shown to be an appropriate representation that allows rendering complex objects in real-time~\cite{grossman1998point}. Recently, point cloud representations were combined with deep learning methods to obtain realistic views even in the presence of noise or sparsity in the point clouds~\cite{aliev2020neural, meshry2019neural, pittaluga2019revealing, wu2020multi}. Other approaches that use a generative neural network for refining the results were also developed for methods that use surfels (oriented planar disks)~\cite{bui2018point} and spheres instead of points~\cite{lassner2021pulsar}. Moreover, the use of deep learning in conjunction with point clouds was facilitated by the development of differentiable rendering models which allow optimizing neural descriptors~\cite{aliev2020neural}, the opacity of points~\cite{kolos2020transpr}, their positions~\cite{wiles2020synsin}, and spheres' radii~\cite{lassner2021pulsar}. 
Our work uses a point-based geometry; however, a generic method for integrating image features was developed instead of optimizing per point descriptors. This extends the single-view feedforward point-based approach to new view synthesis proposed in SynSin~\cite{wiles2020synsin}.

\paragraph{Light field factorization}
In order to handle specular effects, a view-dependent rendering system must be employed. A popular approach used to model such effects is to rely on Spherical Harmonics to represent spherical functions~\cite{sloan2002precomputed}. Spherical Harmonics and Spherical Gaussians have been used to speed up inference of Neural Radience Fields by factorizing the view-dependent appearance~\cite{yu2021plenoctrees}. A similar idea was used in the work of Wizadwongsa et al.~\cite{wizadwongsa2021nex}, where several basis functions were investigated, such as Hemi-spherical harmonics, Fourier Series, and Jacobi Spherical Harmonics with best results achieved by learnable basis functions.
For more detailed information, we refer the reader to a recent overview~\cite{tewari2021advances}.

\section{Method}
\label{sec:method}


Given a set of multiview input images, associated camera parameters, and a point cloud of a static scene, our system generates images from novel views. To do this, neural descriptors, latent vectors representing local geometric and photometric properties, are computed from the information contained in the input views. The latent descriptors are later rasterized using geometric point positions and are converted into final rendering using a refiner (renderer) convolutional network. In contrast to NPBG \cite{aliev2020neural}, our system is learning-based: we predict the neural descriptors instead of optimizing them for each new scene (\Figure{overview}).

In contrast to image-based approaches that find the nearest views from the input set of images to render a novel view, our system creates a single scene model. To construct the model, we process input views in online mode, one at a time, updating the intermediate states of the points in each iteration. These intermediate states are independent of the number of processed views, allowing us to process the scene in constant memory. After processing all input views, we compute final descriptors using the intermediate states. We discuss the modeling stage below, and the rendering process as well as the learning process after that.

\subsection{Modeling 3D Scenes}
\label{sec:preparation_stage}

\paragraph{Feature extraction} The feature extractor, a U-Net-based network \cite{ronneberger2015u}, takes as input an image and outputs a dense feature map (descriptors for each pixel). The feature map has the same spatial size as the input image. The number of channels is equal to the neural descriptor size ($c$=8). We project the points onto the camera canvas and bilinearly sample the descriptors. We use these descriptors to update the intermediate states of the points.

\paragraph{Input image alignment} The bilinearly sampled feature descriptor contains information about the surrounding local patch. The descriptor will change if one rotates the input image since the feature extractor network is not rotation-equivariant by default. We want to bring more consistency across the points' descriptors obtained from different views. For this purpose, we could consider the following options: (i) design the system to be rotation-invariant though this would restrict the representative power of descriptors, (ii) use rotation-equivariant feature extractor \cite{weiler2018learning} though this would inevitably affect the complexity of the feature extractor and/or reduce its capacity, (iii) rotate the input image to a canonical orientation before applying the feature extractor. The last option is the approach we take, i.e., we rotate the input image to align with a \emph{canonical} orientation. We define such orientation so that the world up-axis's projection onto the image plane has a vertical direction (pointing top) in the pixel space (\Figure{overview}-left). We use zero-padding and expand the image size to preserve all image content during the rotation (we change the camera intrinsics and extrinsics to take care of the padding). We call this procedure \emph{input image alignment}.

\paragraph{Estimating point's visibility} To avoid updating descriptors for occluded points, we approximately estimate the visibility of each point. We do this by building a Z-buffer and rasterizing the point cloud onto the image of reduced size $\frac{h}{2^r}{\times}\frac{w}{2^r}$. We consider visible only the points with the minimum Z-value for a pixel location. The visibility reduction factor is set as $r{=}0$ if the point cloud is dense and increased otherwise. In this process, we apply the "nearest" rasterization scheme (\Sec{test_stage}).
While this procedure estimates the visibility approximately, it is fast in contrast to, e.g., constructing the visual hull first like in \cite{katz2007direct}, and does not require to tune the radius of the points as in more advanced rasterization schemes \cite{wiles2020synsin, lassner2021pulsar}.

\paragraph{Aggregation} First, given a new sampled descriptor from the currently processed input view, we want to either set up or update the intermediate states for a point. After all input views are processed, we get the final neural descriptor for the point by exploiting the intermediate states. These two steps completely define the aggregation procedure.


We want to process the incoming descriptors from new input views in online mode: working with one new sample of information at a time and to be memory-independent from the number of input views. Therefore it is not well-suited to consider: (i) Transformer-based \cite{vaswani2017attention} aggregation due to memory limit, (ii) LSTM \cite{hochreiter1997long} and GRU \cite{cho2014learning} recurrent networks as they introduce the undesirable dependence on the order of the input views. Conversely, we design our method to be permutation invariant. One viable option is average or maximum aggregation. However, the drawback of this approach is the loss of information about the view orientations of incoming descriptors, which, as we show, hinders the ability to model view-dependent effects.

Note that view-dependent effects can also be handled by the refiner (renderer) network that can consider view directions during new view synthesis \cite{thies2019deferred,lassner2021pulsar}. However, this requires a more complex and slow refiner network. Alternatively, we choose to make the neural descriptor of each point view-dependent.


\begin{figure}[t]
\centering
\includegraphics[width=0.99\columnwidth]{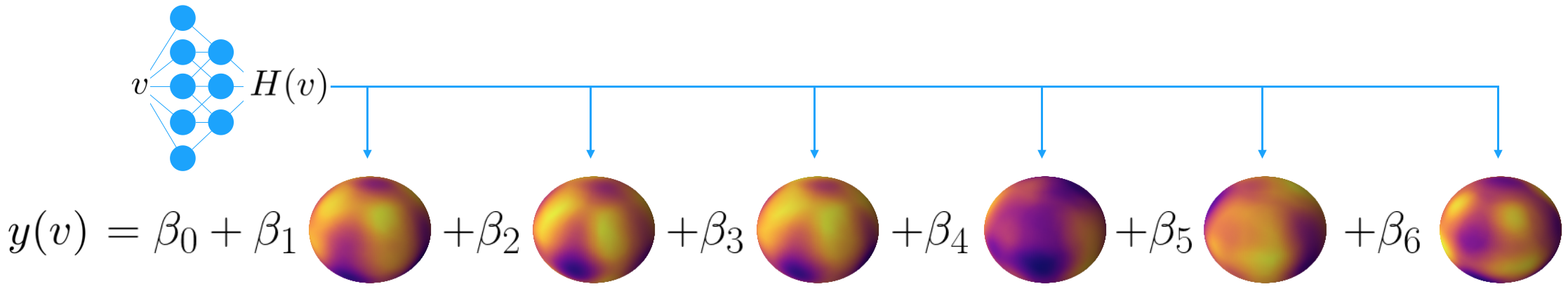}
\figcaption{View-dependent neural descriptor.}{We model the view-dependent neural descriptor $y{:}\ \mathbb{R}^3{\to}\mathbb{R}^c$ as a linear combination of learnable basis functions over the sphere ($H{:}\ \mathbb{R}^3{\to}\mathbb{R}^m$) with coefficients $\beta_i\in\mathbb{R}^c$ (see \Equation{basis} in the main text). For a new scene, given a set of source images, we find $\beta_i$ for each point}
\label{fig:basis}
\end{figure}

In more detail, we model the neural descriptor $y{:}\ \mathbb{R}^3{\to}\mathbb{R}^c$ ($c{=}8$) for the point as a linear combination of learnable basis functions over the sphere:

\begin{equation}
\label{eq:basis}
\underset{\color{blue}1\times c}{y(v)} = \underset{\color{blue}1\times m}{H(v)}\underset{\color{blue}m\times c}{\beta\vphantom{(}}+ \underset{\color{blue}1\times c}{\beta_0\vphantom{(}}
\end{equation}

\noindent where $v$ is an unit length view direction, $H{:}\ \mathbb{R}^3{\to}\mathbb{R}^m$ represents a set of $m$ (we use $m{=}6$) basis functions. $H(v)$ can represent the spherical harmonics (SH) bases, but we model it as the multilayer perceptron (MLP) with weights shared across all points. We found this setting to be superior to SH, see \Sec{experiments}. Coefficients $\beta$ and $\beta_0$ are to be found and are different for each point. See the graphical illustration in \Fig{basis}.

The described approach is similar to NEX \cite{wizadwongsa2021nex} where they model view-dependent RGB values instead of neural descriptors. For each new scene, they optimize $\beta_0$ and an MLP, which takes as input the position of the point and outputs $\beta$. In contrast to NEX~\cite{wizadwongsa2021nex}, to find coefficients $\beta$ and $\beta_0$ for all $N$ points, we solve $N$ multivariate linear regression problems.

For each point, we have a set of pairs $\{(v_k, y_k)\}_{k=1}^K$, where $K$ is the number of input views in which we estimate the point to be visible ($K$ might be different for different points), $v_k\in\mathbb{R}^3$ is a unit-length view direction, $y_k\in\mathbb{R}^c$ is a sampled descriptor from the input image. Given this, we find the parameters of the descriptor as follows:

\begin{equation}
\label{eq:beta_0}
\underset{\color{blue}1\times c}{\beta_0} = \frac{1}{K}\sum_{k=1}^K\underset{\color{blue}1\times c}{y_k}
\end{equation}
\begin{equation*}
\underset{\color{blue}m\times c}{R}\coloneqq\frac{1}{K}\sum_{k=1}^K \underbrace{H(v_k)^Ty_k}_{\color{blue}m\times c}-\frac{1}{K}\sum_{k=1}^K \underbrace{H(v_k)^T\beta_0}_{\color{blue}m\times c}
\end{equation*}
\begin{equation}
\label{eq:beta}
\underset{\color{blue}m\times c}{\beta} = \left(\frac{1}{K}\sum_{k=1}^K \underbrace{H(v_k)^T H(v_k)}_{\color{blue}m\times m}+\frac{\alpha}{K}\underset{\color{blue}m\times m}{I_m\vphantom{\frac{\alpha}{K}}}\right)^{-1}\underset{\color{blue}m\times c}{R}
\end{equation}

\noindent where $I_m$ is the identity matrix, $\beta_0$ captures the mean descriptor, and we set the regularizer $\alpha{=}1$. When a new descriptor sample $y_k$ arrives we update five intermediate states: $K$, $\sum_{k=1}^Ky_k$, $\sum_{k=1}^K H(v_k)^Ty_k$, $\sum_{k=1}^K H(v_k)^T$, $\sum_{k=1}^K H(v_k)^T H(v_k)$. Note that the tensor size of intermediate states does not include the number of input views $K$. For each point, we update its intermediate states until all input views are processed. Then we compute the coefficients $\beta$ and $\beta_0$. We exclude points from the point cloud that were not considered visible in any input view. 

\subsection{Novel view synthesis}
\label{sec:test_stage}

Given the target camera parameters, the final rendering onto the canvas associated with it consists of three steps:

\paragraph{Descriptors calculation step} We calculate descriptors using \Equation{basis}.

\paragraph{Rasterization step} Following NPBG \cite{aliev2020neural}, we construct a pyramid of rasterized \textit{raw} images $\{S_t\}_{t=1}^{T}$ ($T{=}5$ in all our experiments), where each $S_t\in \mathbb{R}^{\left\lfloor\frac{h}{2^{t-1}}\right\rfloor{\times}\left\lfloor\frac{w}{2^{t-1}}\right\rfloor{\times}(c+1)}$ is formed by assigning to every pixel the neural descriptor of the point which passed the depth test, which projects to it under camera full projection transform, and a binary scalar, which indicates a non-empty pixel. The image with the highest resolution in this pyramid provides fine details, while the image with the coarsest resolution suffers the least from surface bleeding and guides the implicit hole filling process inside the refiner network. We fill the pixels that do not have a point projected to them with zeros instead of using a learnable ``void'' descriptor, as it was originally proposed in \cite{aliev2020neural}. In our experiments, this setting leads to better generalization to new views. Alternative rasterization schemes, e.g., soft rasterization (SynSin \cite{wiles2020synsin}) and sphere tracing (Pulsar \cite{lassner2021pulsar}) could be used, still we use the NPBG rasterization due to its speed and lack of additional tunable parameters.

\paragraph{Refinement step} Following NPBG \cite{aliev2020neural}, we process the rasterizations with the refiner network that has a U-Net \cite{ronneberger2015u} architecture with gated convolutions \cite{yu2019free}. The refiner network thus takes as an input the rasterized \textit{raw} image $S_1$ and also appends $S_2, \dots, S_T$ to the input of the corresponding size in intermediate layers of the encoder network. These intermediate inputs guide the network to contend the bleeding surface. The refiner outputs the final RGB image.

\paragraph{Output image alignment} We make a \emph{modification} to the rasterization and refinement steps. Since the refiner network is convolution-based and not rotation-equivariant, the rendered patch will look differently depending on the target camera's y-axis orientation. This is overlooked by previous approaches working with neural descriptors \cite{aliev2020neural, lassner2021pulsar, wiles2020synsin}. Therefore, analogously to the input image alignment (\Sec{training}), we first rasterize and render our final image to the canonical orientation and then rotate the produced image to align with the original orientation (\Figure{overview}-right).

\subsection{Training}
\label{sec:training}

We select the batch of target views from different scenes and randomly crop patches of the same size during each training iteration. We use patches instead of entire images to reduce memory load during training. For each patch, during training only, we select three relevant views from which we aggregate descriptors and then render the final image.

\paragraph{View selection} To select the relevant views, we follow the approach from MVSNet \cite{yao2018mvsnet}: we calculate a score for each input view by taking a sum of scores of points that are visible both in the target patch and the input view. We stress that view selection is used only to improve training and is not used after training, i.e., during scene fitting or new view synthesis. 
The point score is angle-based: the smaller is the angle between the tracks connecting two camera centers and the point, the higher the score. For the exact formula, we refer the reader to \cite{yao2018mvsnet}. The input view scores define the probabilities of a discrete multinomial distribution from which we sample the indices of relevant input views. We further modify the original procedure as follows. As we want to avoid the case when all selected images are similar and do not cover some part of the target image, we do not sample all relevant views at once but do that sequentially. After selecting the first view, we remove the points visible in this image and recalculate the scores for the remaining views. We repeat the selection and removal until we get the desired number of relevant views.

\paragraph{Cropping} To avoid running the feature extractor network on the selected images at the original large size, we crop them such that the crops contain as many points from the target patch as possible.

\paragraph{Loss} We employ the following train loss:
$$
\loss{}(I,I_{\text{gt}})=\lambda_{1}\loss{\text{vgg}}(I,I_{\text{gt}})+\lambda_{2}\loss{1}(I\downarrow_4,I_{\text{gt}}\downarrow_4)+\lambda_{3}\loss{\text{reg}}(I_{\text{gt}})
$$
\noindent where $I$ is a rendered image, $I_{\text{gt}}$ is a reference ground-truth image. Similar to \cite{aliev2020neural, sevastopolsky2020relightable} $\loss{\text{vgg}}$ is a VGG-19 \cite{Simonyan15} perceptual loss and $\loss{1}$ loss is used to match four times bilinearly down-sampled versions of the images to prevent high-frequency detail smoothing while encouraging color preservation. Additionally, we introduce a self-supervised regularization loss $\loss{\text{reg}}$ that improves the learning signal for the refiner network $R$ by providing high-quality descriptors extracted directly from the ground truth image (unavailable at test time).

$\loss{\text{reg}}(I_{\text{gt}}){=}\loss{1}(R(\text{pyr}(sg(F(I_{\text{gt}})))), I_{\text{gt}})$, where $F$ is a feature extractor, $\text{pyr}(\cdot)$ takes as input the dense output from the feature extractor $F$ and outputs a pyramid of images. The first image in the pyramid is $F(I_\text{gt})\in\mathbb{R}^{h\times w\times c}$. Each new level image is obtained using $2{\times}2$ average pooling of the previous one. $sg$ (stop-grad) is the non-differentiable version of the identity function and is used to avoid shortcut solutions e.g. by directly passing the original image through the network. We set $\lambda_{1}{=}1$, $\lambda_{2}{=}2500, \lambda_{3}{=}1000$.
\begin{table*}[ht]
\centering
\resizebox{\textwidth}{!}{
\begin{tabular}{lccccccccccccc}
\toprule
& & \multicolumn{3}{c}{Nerf-Synthetic} & \multicolumn{3}{c}{ScanNet} & \multicolumn{3}{c}{DTU} & \multicolumn{3}{c}{H3DS} \\ \cmidrule(lr){3-5}\cmidrule(lr){6-8}\cmidrule(lr){9-11}\cmidrule(lr){12-14}
Method & \begin{tabular}[c]{@{}l@{}}Per scene\\ optimization\end{tabular} & \textbf{PSNR}$\uparrow$      & \textbf{SSIM}$\uparrow$      & \textbf{LPIPS}$\downarrow$      & \textbf{PSNR}$\uparrow$      & \textbf{SSIM}$\uparrow$      & \textbf{LPIPS}$\downarrow$      & \textbf{PSNR}$\uparrow$      & \textbf{SSIM}$\uparrow$      & \textbf{LPIPS}$\downarrow$      & \textbf{PSNR}$\uparrow$      & \textbf{SSIM}$\uparrow$      & \textbf{LPIPS}$\downarrow$      \\ \midrule
SVS\cite{riegler2021stable}
 & \xmark        &
22.81       & 0.919     & \underline{0.104}         & 
\underline{23.32}       & \textbf{0.771}     & \textbf{0.445}         & 
20.98       & 0.897     & \underline{0.162}         & 
18.96       & \underline{0.798}     & \underline{0.210}         \\ 
IBRNet\cite{wang2021ibrnet}
 & \xmark        &
\textbf{29.47}       & \textbf{0.955}     & 0.157         & 
\textbf{23.34}       & 0.760     & \underline{0.494}         & 
\textbf{25.81}       & \textbf{0.924}     & 0.231         & 
\underline{20.30}       & 0.791     & 0.279         \\ 
\textbf{NPBG++ (Ours)} & \xmark &
\underline{26.06}       & \underline{0.936}     & \textbf{0.071}         & 
23.11       & \underline{0.766}     & 0.502         & 
\underline{23.23}       & \underline{0.915}     & \textbf{0.154}         & 
\textbf{21.80}       & \textbf{0.818}     & \textbf{0.177}         \\ 
\cmidrule{1-14}
NPBG\cite{aliev2020neural}
 & \cmark     & 
28.62       & 0.946     & 0.058         & 
25.09       & 0.737     & \underline{0.459}         & 
26.00       & 0.913     & \underline{0.125}         & 
\underline{24.68}       & 0.827     & \underline{0.146}         \\ 
NeRF\cite{mildenhall2020nerf}
   & \cmark    & 
\underline{32.49}       & \underline{0.970}     & \textbf{0.041}         &
\textbf{25.74}       & \textbf{0.780}     & 0.537         & 
\textbf{26.92}       & 0.913     & 0.198         & 
23.88       & 0.833     & 0.178         \\ 
SVS$_{\emph{ft}}$\cite{riegler2021stable}
   & \cmark    & 
23.37       & 0.919     & 0.101         & 
22.31       & 0.610     & 0.543         & 
20.72       & 0.864     & 0.190         & 
20.12       & 0.770     & 0.197         \\ 
IBRNet$_{\emph{ft}}$\cite{wang2021ibrnet}   & \cmark    & 
\textbf{32.51}       & \textbf{0.972}     & 0.144         & 
24.42       & \underline{0.774}     & 0.493         & 
23.80       & 0.917     & 0.222         & 
\underline{24.68}       & \textbf{0.850}     & 0.195         \\
\textbf{NPBG++$_{\emph{ft-system}}$ (Ours)}   & \cmark    & 
26.24       & 0.940     & 0.064         & 
23.48       & 0.768     & 0.490         & 
24.05       & \underline{0.919}     & 0.147         & 
23.79       & 0.836     & 0.155         \\ 
\textbf{NPBG++$_{\emph{ft}}$ (Ours)}   & \cmark    & 
28.67       & 0.952     & \underline{0.050}         & 
\underline{25.27}       & 0.772     & \textbf{0.448}         & 
\underline{26.08}       & \textbf{0.928}     & \textbf{0.123}         &
\textbf{24.91}       & \underline{0.845}     & \textbf{0.137}         \\
\bottomrule
\end{tabular}
}
\figcaption{Quantitative evaluations.}{For each dataset, we compute the metrics \cite{zhang2018perceptual} on holdout frames averaged across holdout scenes. Subscript \emph{ft} indicates finetuned versions of the methods. In the case of NPBG++$_{\emph{ft}}$ we directly finetune coefficients ($\beta$, $\beta_0$) and the refiner. In the case of NPBG++$_{\emph{ft-system}}$ we finetune the feature extractor, aggregator (MLP: neural basis functions), and refiner.}
\label{tab:comparisons_short}
\end{table*}
\section{Experimental Results}
\label{sec:experiments}


\paragraph{Datasets} We validate the effectiveness of the proposed method on four different datasets: ScanNet \cite{dai2017scannet}, NeRF-Synthetic \cite{mildenhall2020nerf}, H3DS \cite{ramon2021h3d}, and DTU \cite{jensen2014large}. For H3DS and DTU, we apply masks to all images to make the background white. To obtain the point-cloud geometry in the case of ScanNet, we use the depth data in the same manner as NPBG \cite{aliev2020neural}. For Nerf-synthetic, DTU, and H3DS, we use the multi-view stereo approach PatchmatchNet~\cite{wang2020patchmatchnet} which offers a good balance between speed and point cloud quality. We have found it to be fast and yield results with more surface coverage compared to other MVS approaches such as COLMAP \cite{schoenberger2016vote}. For further details about the datasets, we refer the reader to the Supplementary materials.

\paragraph{Training Details} We consider only four different scenes in each epoch to accelerate training time by caching the input point clouds. Each iteration, we sample eight target patches of size 256x256. Each epoch lasts 2000 iterations. To avoid overfitting, we apply color augmentations for the target patch and input view patches. We train our system (the feature extractor, aggregator, and refiner) on the ScanNet dataset. We start from the ScanNet pretraining for all other datasets and fine-tune the system on the training scenes for that dataset. For implementation details we refer the reader to the Supplementary materials.

\paragraph{Compared methods} We compare our results with several state-of-the-art neural rendering algorithms.

\begin{itemize}
    \item \textbf{NPBG \cite{aliev2020neural}:} a neural-point-based graphics approach aimed at per-scene optimization. We reimplement the original network in a faithful manner and pretrain the refiner network on the same set of scenes as our pipeline to provide a fair comparison.
    
    \item \textbf{NeRF \cite{mildenhall2020nerf}}: a volume-rendering-based approach aimed at per-scene optimization. We use the slightly faster PyTorch implementation \cite{lin2020nerfpytorch}.
    
    \item \textbf{SVS \cite{riegler2021stable}}: Stable View Synthesis uses a geometric scaffold on the surface of which it aggregates image features. The method is capable of rendering new scenes without optimization; further fine-tuning of the system is optional.
    
    \item \textbf{IBRNet \cite{wang2021ibrnet}:} an image-based rendering approach that learns a generic view interpolation function applicable to novel scenes.
\end{itemize}

\begin{table}[t]
\centering
\resizebox{0.99\columnwidth}{!}{
\begin{tabular}{lccc}
\toprule
Method & H3DS & DTU & Nerf-Synthetic \\
\midrule
NPBG++ w/ H3DS ft         &
 \textbf{0.177}                 & 
 0.176         & 
0.102         \\ 
NPBG++ w/ DTU ft       &
 \underline{0.209}         & 
 \textbf{0.154}         & 
 \underline{0.093}        \\ 
NPBG++ w/ Nerf ft &
 0.212               & 
 \underline{0.164}         & 
\textbf{0.071}         \\ 
\bottomrule
\end{tabular}
}
\figcaption{Cross-dataset generalization.}{We report LPIPS$\downarrow$ on holdout frames averaged across holdout scenes for each dataset.}
\label{tab:generalization}
\end{table}
\begin{figure*}[ht]
\centering
\includegraphics[width=0.99\textwidth]{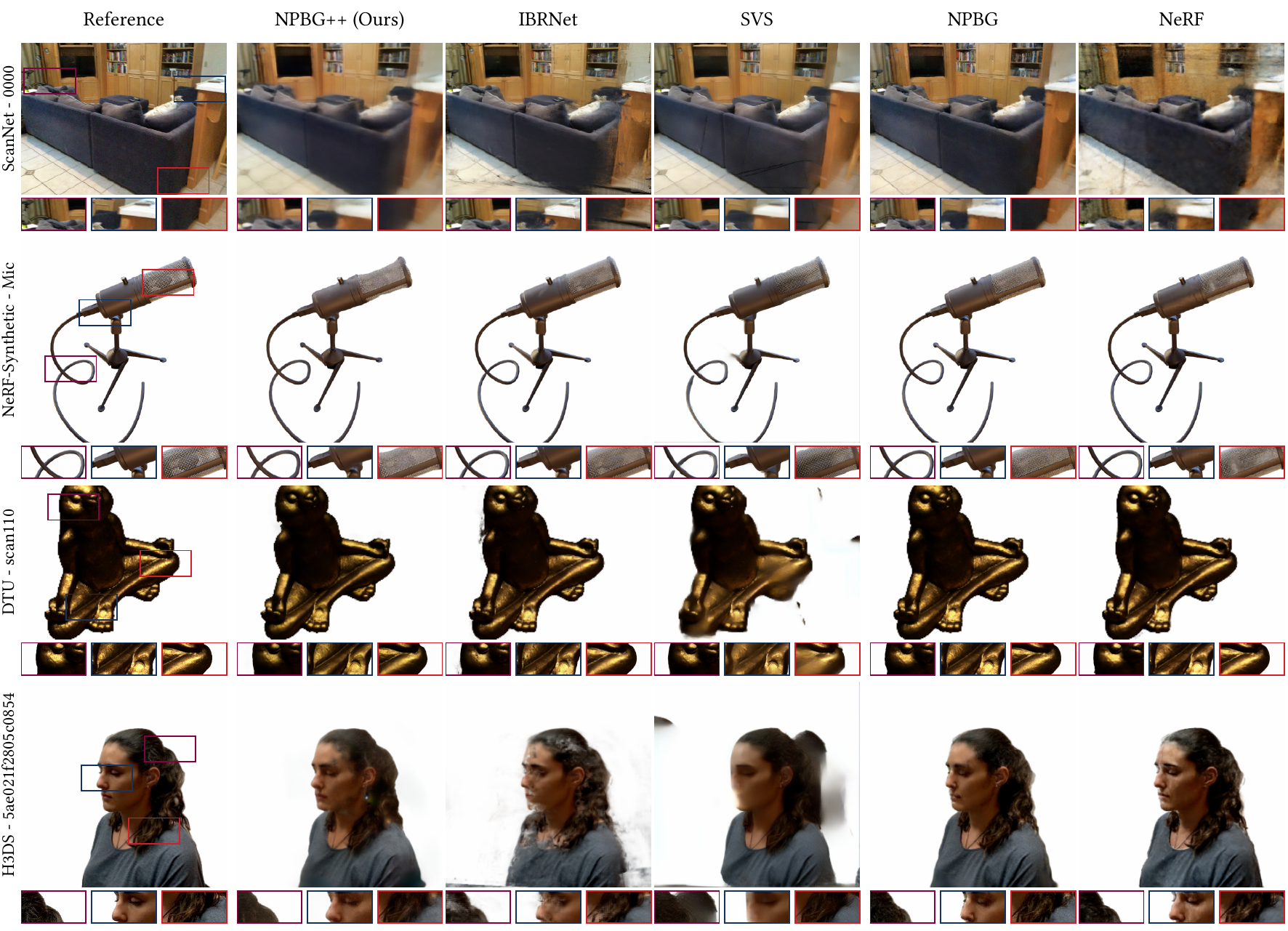}
\figcaption{Qualitative evaluations.}{Comparisons with otimization-based approaches (NPBG\cite{aliev2020neural}, NeRF\cite{mildenhall2020nerf}) and learning based approaches (IBRNet\cite{wang2021ibrnet}, SVS\cite{riegler2021stable}) on ScanNet\cite{dai2017scannet}, NeRF-Synthetic\cite{mildenhall2020nerf}, DTU\cite{jensen2014large}, H3DS\cite{ramon2021h3d} scenes.}
\label{fig:comparisons}
\end{figure*}

\paragraph{Evaluation} \Table{comparisons_short} shows the quantitative comparisons with the state-of-the-art on several standard metrics (SSIM, PSNR, and LPIPS \cite{zhang2018perceptual}). \Figure{comparisons} shows the qualitative comparisons. \Table{generalization} demonstrates the generalization ability of our method.




The metrics for results obtained without per-scene optimization show that our method can produce superior renderings to SVS and, generally, on-par with IBRNet - the state-of-the-art NVS method with fast generalization to new scenes. While in some metrics, IBRNet is better on certain scenes, the rendering speed advantage of NPBG++ over IBRNet is drastic ($\approx 1000 \times$). 

The numerical results in the fine-tuning case affirm that our view-dependence modeling is effective, allowing us to outperform NPBG on all datasets. The proposed approach obtains leading scores on DTU and H3DS scenes and is a close competitor on ScanNet and NeRF-Synthetic datasets. On the latter, our method is limited by the lower quality geometry estimated by the MVS system and could potentially be improved by using a rasterization scheme differentiable w.r.t to points' positions \cite{wiles2020synsin, lassner2021pulsar}.

The qualitative comparison reveals that our method can generate marginally blurry but comprehensive and consistent views of the scenes. Images rendered by IBRNet can introduce artifacts when parts of the image are not included in enough source images (e.g.~on \texttt{ScanNet{-}0000}). This may happen if there are only a few training images or if the view selection method underperforms. Note that all methods from the official implementation were tested, and the reported images and scores are obtained with the best performing one. NPBG++ does not suffer from this problem because it aggregates features from all images. SVS displays difficulties in representing thin structures (mic - NeRF synthetic) and often produces spurious artifacts in object-centric scenes. NPBG generally yields high-quality results; however, it cannot handle view-dependent effects by design, leading to bland results for shiny surfaces. NeRF offers generally good results, but the high fitting and rendering time make it impractical for many use-cases.

\paragraph{Runtime analysis} For inference, there are two different stages for which we compare the speed of several state-of-the-art methods. In the first (fitting) stage, the algorithms capture information from the source images. For methods based on per-scene optimization, this generally means training the neural representation of the scene. For IBRNet, it means running the feature extractor on the selected neighboring views. For our method, it refers to the 3D modeling stage (\Sec{preparation_stage}). Approaches that use geometric proxies, such as SVS, NPBG, and NPBG++, need to construct the 3D representation as part of this stage. The time required for this process is included in time measurements to provide a fair comparison.
This first step is performed a single time per scene, after which each model should be able to render any number of novel views, i.e., the second stage (rendering). As it can be observed in \Figure{teaser}, NeRF and IBRNet have very high rendering times. Notably, the time needed to render a novel image using IBRNet is larger than the entire fitting process of our method. NeRF, PlenOctrees, and NPBG require per-scene fitting, which leads to larger times until good quality results are obtained. Out of the methods which do not require optimization, our model has the smallest time-to-render (fitting + rendering) time: SVS is slowed down by surface estimation, and IBRNet's timing is dominated by the rendering step.

\begin{figure}[t]
\centering
\includegraphics[width=0.99\columnwidth]{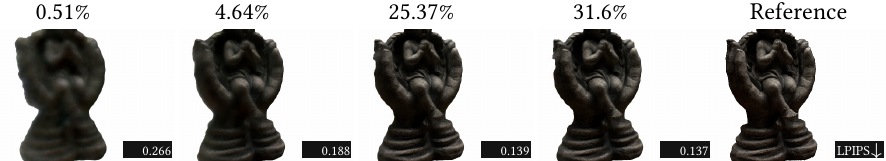}
\figcaption{Robustness with respect to the point cloud sparsity.}{We randomly drop different amounts of points from the point cloud. In each case, we show the percentage of filled pixels after rasterization. The effect is illustrated on scan118 from DTU.}
\label{fig:num_points_impact}
\end{figure}

\begin{figure}[t]
\centering
\includegraphics[width=0.99\columnwidth]{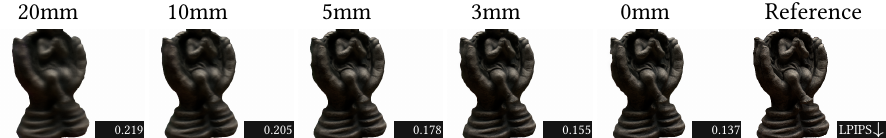}
\figcaption{Robustness with respect to geometry noise.}{We move every point along a random unit direction by a specified value for each case. We report LPIPS$\downarrow$ averaged across holdout frames. The effect is illustrated on scan118 from DTU.}
\label{fig:noise_impact}
\end{figure}

\begin{figure}[t]
\centering
\includegraphics[width=0.99\columnwidth]{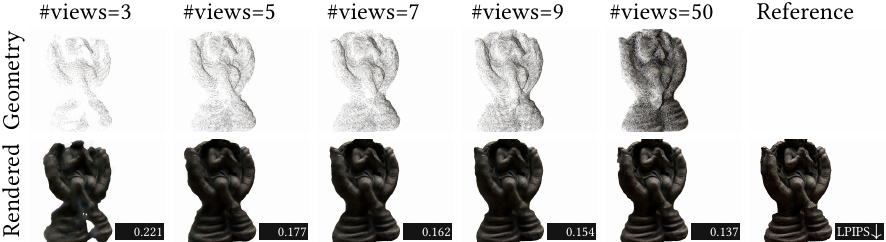}
\figcaption{Robustness with respect to the number of input views.}{We report LPIPS$\downarrow$ averaged across holdout frames. The effect is illustrated on scan118 from DTU. \vspace{-3mm}}
\label{fig:num_views_impact}
\end{figure}

\paragraph{Ablations} First of all, we demonstrate the impact of input and output image alignment. To demonstrate it, we run different modifications of our system and NPBG \cite{aliev2020neural} (\Table{alignment}). Adding output image alignment to NPBG improves metrics. In the case of NPBG++, using either only input or only output image alignment makes metrics worse. This is likely caused by the misalignment between distributions of the camera's y-axis orientations in input and output images. However, turning on both input and output alignments tackles this problem and improves metrics. Second, we test the robustness of our system to geometry imperfections, such as the sparsity of the point cloud (\Figure{num_points_impact}), noise (\Figure{noise_impact}) and also few images scenario (\Figure{num_views_impact}). The method shows graceful degradation in all cases.
Third, we compare different variants for the aggregation procedure (\Table{aggregation}). We observe that the view-dependent variants outperform the simple average aggregation. In the case of spherical harmonics (SH), we compare only with the 4 harmonics setting due to memory constraints.
The best results are obtained using the MLP with $m{=}6$ basis functions.

\begin{table}[t]
\centering
\resizebox{0.8\columnwidth}{!}{
\begin{tabular}{lccc}
\toprule
 & \textbf{PSNR}$\uparrow$ & \textbf{SSIM}$\uparrow$ & \textbf{LPIPS}$\downarrow$ \\
\midrule
average & 21.41 & 0.749 & 0.525 \\
spherical harmonics (m=4) & 22.19 & 0.753 & 0.513 \\
mlp (m=3) & 22.97 & 0.756 & \textbf{0.500} \\
mlp (m=6) (default) & \textbf{23.11} & \textbf{0.766} & \underline{0.501} \\
\bottomrule
\end{tabular}
}
\figcaption{Aggregation.}{We test different options for the aggregation procedure as discussed in \Sec{preparation_stage}. We report metrics averaged across three holdout ScanNet\{0,43,45\} scenes.}
\label{tab:aggregation}
\end{table}
\begin{table}[t]
\centering
\resizebox{0.99\columnwidth}{!}{
\begin{tabular}{lccccc}
\toprule
Method & \begin{tabular}[c]{@{}c@{}}Input Image\\ Alignment\end{tabular} & \begin{tabular}[c]{@{}c@{}}Output Image\\ Alignment\end{tabular} & \textbf{PSNR}$\uparrow$ & \textbf{SSIM}$\uparrow$ & \textbf{LPIPS}$\downarrow$ \\
\midrule
NPBG (original) \cite{aliev2020neural} & n/a & \xmark & 26.00 & 0.913 & \textbf{0.125} \\
NPBG  & n/a & \cmark & \textbf{26.16} & \textbf{0.920} & \underline{0.126} \\
\midrule
NPBG++ & \xmark & \xmark & 22.53 & 0.912 & \textbf{0.149} \\
NPBG++ & \cmark & \xmark & 21.87 & 0.876 & 0.201 \\
NPBG++ & \xmark & \cmark & 22.47 & 0.879 & 0.203 \\
NPBG++ (default) & \cmark & \cmark & \textbf{23.23} & \textbf{0.915} & \underline{0.154} \\
\bottomrule
\end{tabular}
}
\figcaption{The impact of alignment.}{We provide empirical evidence on how the input and output image alignment (\Sec{preparation_stage}, \Sec{test_stage}, \Figure{teaser}) improve the final result for neural point-based graphics pipelines. We report metrics averaged across three holdout DTU scenes\{110,114,118\}. In the case of NPBG, a separate network was trained for each scene and each alignment.}
\label{tab:alignment}
\end{table}
\section{Summary and Limitations}
\label{sec:conclusion}

In this paper, we introduced NPBG++, a new system for novel view synthesis that can rapidly obtain the neural representation of a scene, passing a single time through all available source views. We obtain the coefficients for learned basis functions by solving online linear regression that allows us to model view-dependent neural descriptors. These descriptors are then used to render novel views from arbitrary camera poses easily. We showed that our system can produce good quality results at a fraction of the time required for other methods while also achieving real-time rendering for medium resolution images.

NPBG++ inherits some limitations from NPBG in that it still requires an explicit underlying geometric model (point cloud).
In the case of highly erroneous point clouds or inaccurate camera alignments, as well as extreme scale changes, renderings can suffer from blurriness.
We additionally introduced an $\alpha$ hyperparameter in \Equation{beta} but keeping the same value ($\alpha{=}1.0$) for all points may over-regularize the solution and dampen view-dependent effects.

\paragraph{Acknowledgment} The Authors acknowledge the use of computational resources of the Skoltech CDISE supercomputer Zhores \cite{zacharov2019zhores} for obtaining the results presented in this paper. The work was supported by the Analytical center under the RF Government (subsidy agreement 000000D730321P5Q0002, Grant No. 70-2021-00145 02.11.2021).





{
    \clearpage
    \small
    \bibliographystyle{ieee_fullname}
    \bibliography{macros,main}
}


\end{document}